\documentclass[letterpaper, 10 pt, conference]{ieeeconf}  

\IEEEoverridecommandlockouts                              

\usepackage[english]{babel}
\usepackage{stfloats}
\usepackage{graphicx}
\usepackage{epstopdf}
\usepackage{subfigure}

\usepackage{amsthm}

\graphicspath{{figures/}}
\usepackage{makecell}
\usepackage{sidecap}
\usepackage{amsmath}
\usepackage{multirow}
\usepackage{amssymb}
\usepackage{amsmath}
\usepackage{bm}
\newsavebox\CBox

\usepackage{amsthm,amsmath,amssymb}
\usepackage{mathrsfs}

\usepackage{siunitx}
\usepackage{hyperref}

\usepackage{graphicx}
\usepackage{subfigure}
\usepackage{siunitx}

\usepackage[ruled]{algorithm2e}

\usepackage{epstopdf}
\usepackage{booktabs}
\usepackage{cite,soul}

\theoremstyle{definition}

\addto\captionsenglish{}

\usepackage{titlesec}

\title{\LARGE \bf
Learning the References of Online Model Predictive Control \\for Urban Self-Driving }

\author{Yubin Wang, Zengqi Peng, Yusen Xie, Yulin Li, Hakim Ghazzai, and Jun Ma
\thanks{Yubin Wang, Zengqi Peng, Yusen Xie, Yulin Li, and Jun Ma are with the Robotics and Autonomous Systems Thrust, The Hong Kong University of Science and Technology (Guangzhou), Guangzhou, China (email: ywang575@connect.hkust-gz.edu.cn; zpeng940@connect.hkust-gz.edu.cn; yxie827@connect.hkust-gz.edu.cn; yline@connect.ust.hk; jun.ma@ust.hk)
}
\thanks{Hakim Ghazzai is with the Division of Computer, Electrical and Mathematical Science and Engineering, King Abdullah University of Science and Technology, Thuwal, Saudi Arabia 
(email: Hakim.Ghazzai@kaust.edu.sa)
}
}

\begin{document}

\maketitle
\thispagestyle{empty}
\pagestyle{empty}


\begin{abstract}
In this work, we propose a novel learning-based model predictive control (MPC) framework for motion planning and control of urban self-driving.
\textcolor{black}{In this framework, instantaneous references and cost functions of online MPC are learned from raw sensor data without relying on any oracle or predicted states of traffic. Moreover, driving safety conditions are 
latently encoded via the introduction of a learnable instantaneous reference vector.}
\textcolor{black}{In particular, we implement a deep reinforcement learning (DRL) framework for policy search, where practical and lightweight raw observations are processed to reason about the traffic and provide the online MPC with instantaneous references.}
The proposed approach is \textcolor{black}{validated in a high-fidelity simulator
, where our development manifests remarkable adaptiveness to complex and dynamic traffic.} 
\textcolor{black}{Furthermore, sim-to-real deployments are also conducted to evaluate the generalizability of the proposed framework in various real-world applications.} 
\textcolor{black}{Also, we provide the open-source code and video demonstrations at the project website:} \href{https://latent-mpc.github.io/}{\textit{https://latent-mpc.github.io/}.}

\end{abstract}

\section{Introduction}
Safe and efficient driving strategies are incredibly essential for the wider adoption of self-driving technology in urban and residential scenarios,
Nevertheless, the relationship between the strong safety and high driving efficiency is a trade-off, as conservative maneuvers could sacrifice efficiency while aggressive driving strategies lack the safety guarantee. 
Hence, it is crucial to develop advanced motion planning and control strategies for self-driving vehicles to ensure the safety conditions, perform agile maneuvers, and manifest improved efficiency.
Optimization-based methods, particularly model predictive control (MPC), have been widely studied due to their capability to optimize the feasible trajectory respecting various constraints. Essentially, the urban self-driving problem can be formulated as an optimization problem with nonlinear vehicle dynamics and various constraints of other types \cite{eiras2021two, schwarting2017safe,adajania2022multi}. However, the performance attained by such methods is degraded due to the conservative motions and undesirable driving behaviors, 
especially in traffic environments with high complexity and dynamicity.
On the other hand, as a representative reinforcement learning (RL) method, deep reinforcement learning (DRL) aims to learn a neural network policy that can map high-dimensional raw observation features directly to control commands
\cite{fuchs2021super, song2021autonomous}. Apparently, such RL-based methods demonstrate the advantages of forgoing the need for dynamics modeling and online optimization~\cite{fuchs2021super}. However, the instability and poor generalizability of the learned policy are the commonly encountered issues that hinder their applications.
Furthermore, a promising way that facilitates effective and reliable driving strategies lies in the design of a hybrid framework that bridges MPC and DRL, in which the neural policy produces decision variables as references to formulate cost terms of the optimization problem in the MPC scheme. 
Thus, it facilitates improved maneuverability of the vehicle, and also ensures the feasibility as well as the reliability of the generated trajectories.

\begin{figure}[!t] 
    \centering   
    \includegraphics[trim=0 0 0 0, width=1.0 \linewidth]{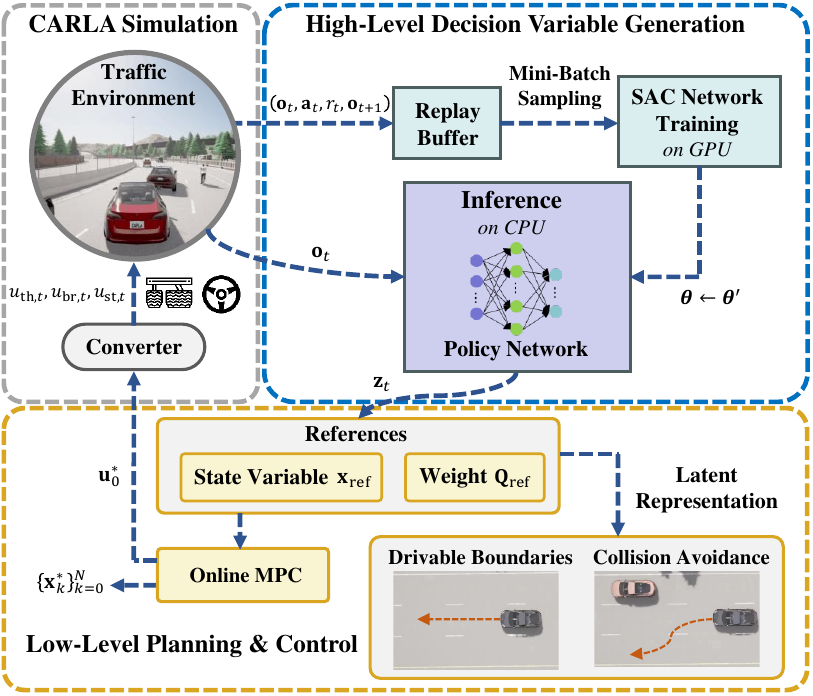}
   \caption{Overview of the proposed framework for urban self-driving. A policy network is trained to produce instantaneous decision variables for low-level online MPC, whose cost functions are modulated to latently \textcolor{black}{encode the safety conditions of collision avoidance and drivable surface boundaries.} }
    \label{overview}
    \vspace{-0.4cm}
\end{figure}

In this paper, we propose a novel learning-based MPC framework for urban self-driving, where motion safety are latently encoded by modulating the cost functions of MPC without relying on any oracle or predicted states of traffic. 
In this sense, the proposed approach learns from raw onboard sensor data to formulate the embodied MPC and it is tolerant to various uncertainties in the measurements.
Also, the policy search for the real-time generation of decision variables is cast to a DRL problem, where the policy network produces the step actions as instantaneous references for MPC. Here, the soft actor-critic (SAC) algorithm is employed in the proposed framework to update the policy, where training data is uniformly sampled from the replay buffer. The overview of the proposed framework is depicted in Fig. \ref{overview}.
The main contributions of this work are summarized are as follows:
    A novel learning-based MPC framework is proposed without prior knowledge about states of traffic, and instantaneous references are learned from raw onboard sensor data to latently encode the driving safety.
    {The learning of pertinent decision variables is formulated as policy search, where an RL agent is tasked to reason about the traffic and produce decision variables with visualized interpretability of driving safety.}
    The performance of proposed framework is evaluated both in a high-fidelity simulator and through sim-to-real validations. Improved driving efficiency and safety as well as alleviated computational burden of solving the optimization problem are demonstrated as compared with baseline methods.
    



\section{Related Works}
\label{sec:related}
Learning-based MPC for motion planning and control is an emerging technique in the area of self-driving and other types of robotic applications, where appropriate learning techniques can be incorporated to model or parameterize the critical factors of MPC. 
A typical approach in learning-based MPC aims to model complex systems with classical machine learning or deep learning techniques. Gaussian process (GP) can be utilized to represent the model error to improve the simple nominal vehicle dynamics model online~\cite{kabzan2019learning}, or approximate the nonlinearities of the dynamics in the presence of uncertainty \cite{hewing2019cautious}, where the solution quality of MPC is improved and the computational burden is alleviated. 
On the other hand, deep neural networks can also be applied for accurate predictions of the system dynamics, where knowledge-based neural ordinary differential equations are adopted to account for residual and uncertain dynamics for quadrotors in \cite{chee2022knode}. 
Additionally, in \cite{lenz2015deepmpc}, a deep neural network architecture is utilized to represent the variant dynamics model for robotic tasks in an online MPC framework, where the MPC is empowered to adaptively tune the dynamics for variant tasks. Moreover, a differentiable constrained optimizer is proposed and embedded as a layer in neural networks to learn behavioural inputs in \cite{shrestha2023end}, where the performance of MPC-based planning is improved towards dense traffic.


The other paradigm of learning-based MPC is to utilize RL techniques to formulate the cost functions of MPC. 
In \cite{lowrey2018plan} and \cite{karnchanachari2020practical}, value learning is utilized to approximate the costs of MPC, which allows for better policy quality beyond local solutions within a reduced planning horizon.
In \cite{zarrouki2021weights}, the context-dependent optimal weights are learned automatically and adapted online for the cost functions of MPC. 
Furthermore, RL can also be integrated into the MPC framework to learn the high-level decision variables for cost formulation. 
In \cite{song2020learning} and \cite{song2022policy}, a high-level policy with the representation of Gaussian distribution is proposed, with which the traversal time of flying through a swinging gate for a quadrotor is determined. 
Furthermore, SE(3) decision variables are learned as state references of the MPC in~\cite{wang2023learning}, with which the quadrotor can traverse a moving and rotating gate. 
In \cite{wang2023chance}, the augmented decision variables are introduced to parameterize the cost functions of high-level MPC for the task of chance-aware lane change in dense traffic environments. 
Despite the success of producing desired decision variables for MPC in \cite{song2020learning, song2022policy, wang2023learning, wang2023chance}, the high-level policy is trained by episodic RL through evaluating the quality of whole trajectories. 
Essentially, the delayed feedback and the difficulty in credit assignment degrade the performance of episodic RL in long-term decision-making. This is because it is challenging to determine which actions should be responsible for the final outcomes, and the agent cannot adapt its behavior at specific steps based on cumulative and delayed reward signals.
Therefore, formulating the generation of decision variables as an episodic RL problem could potentially hinder the maneuverability and safety of vehicles in highly dynamic and complex traffic environments.
To address this issue, we aim to incorporate the step-based RL for policy search, where the actions evolve with intermediate reward signals. Furthermore, we formulate the real-time generation of decision variables as a sequential decision-making problem.
In this sense, the proposed framework is able to demonstrate improved adaptiveness to the high complexity and dynamicity of traffic in urban self-driving tasks.

\section{Preliminaries and Problem Statement}
\label{sec:statement}

 
In this work, the bicycle model is adopted to describe the vehicle's kinematics. The state vector of the vehicle in the global coordinate $\mathcal{W}_g$ is defined as 
$\mathbf{X}= \begin{bmatrix}  X & Y & \Psi & V \end{bmatrix} ^{\top}$,
where $X$ and $Y$ denote the X-coordinate and Y-coordinate position of the center of the vehicle, $\Psi$ is the heading angle, and $V$ is the speed. Also, we integrate the control inputs into a vector as $\mathbf{u}= \begin{bmatrix} a & \delta \end{bmatrix} ^{\top}$, where $a$ and $\delta$ are the acceleration and steering angle. 
Subsequently, by modeling the vehicle as a rectangle, the nonlinear kinematic model of the vehicle in continuous time is given by:
    \begin{equation}
    \dot{\mathbf{X}}=f(\mathbf{X}, \mathbf{u})=
    \left[\begin{array}{c}
    V \cos (\Psi+\delta) \\
    V \sin (\Psi+\delta) \\
    \frac{2V}{L} \sin \delta \\
    a 
    \end{array}\right],
    \label{kin}
    \end{equation}
where $L$ is the inter-axle distance of the vehicle.

For the convenience of formulating the optimization problem in general urban self-driving scenarios, we exploit a transformation of the vehicle state vector $\mathbf{X}$ from the global coordinate $\mathcal{W}_g$ into the road centerline reference coordinate $\mathcal{W}_\mathrm{ref}$.
It is assumed that 
the two-dimensional centerline of the road $\mathcal{P}_\mathrm{ref}$ is detected and the length of centerline $|\mathcal{P}_\mathrm{ref}|$ is parameterized by the longitudinal distance $\lambda$ from its start, the point on the centerline can be defined as
$(X^{\mathcal{P}_\mathrm{ref}}(\lambda), Y^{\mathcal{P}_\mathrm{ref}}(\lambda))$, where $\lambda \in [0, |\mathcal{P}_\mathrm{ref}|]$.
The tangential and normal vectors of the centerline in the global coordinate $\mathcal{W}_g$, denoted by $\mathbf{t}_\lambda$ and $\mathbf{n}_\lambda$, can be written as:
\begin{equation}
\mathbf{t}_\lambda=\left[\begin{array}{c}
\frac{\partial X^{\mathcal{P}_{\mathrm{ref}}}(\lambda)}{\partial \lambda} \\
\frac{\partial Y^{P_{\mathrm{ref}}(\lambda)}}{\partial \lambda}
\end{array}\right], \quad \mathbf{n}_\lambda=\left[\begin{array}{c}
\frac{-\partial Y^{\mathcal{P}_{\mathrm{ref}}}(\lambda)}{\partial \lambda} \\
\frac{\partial X^{\mathcal{P}_{\mathrm{ref}}}(\lambda)}{\partial \lambda}
\end{array}\right] .
\end{equation}
\begin{figure}[!t] 
    \centering   
    \includegraphics[trim=0 0 0 0, width=1.0\linewidth]{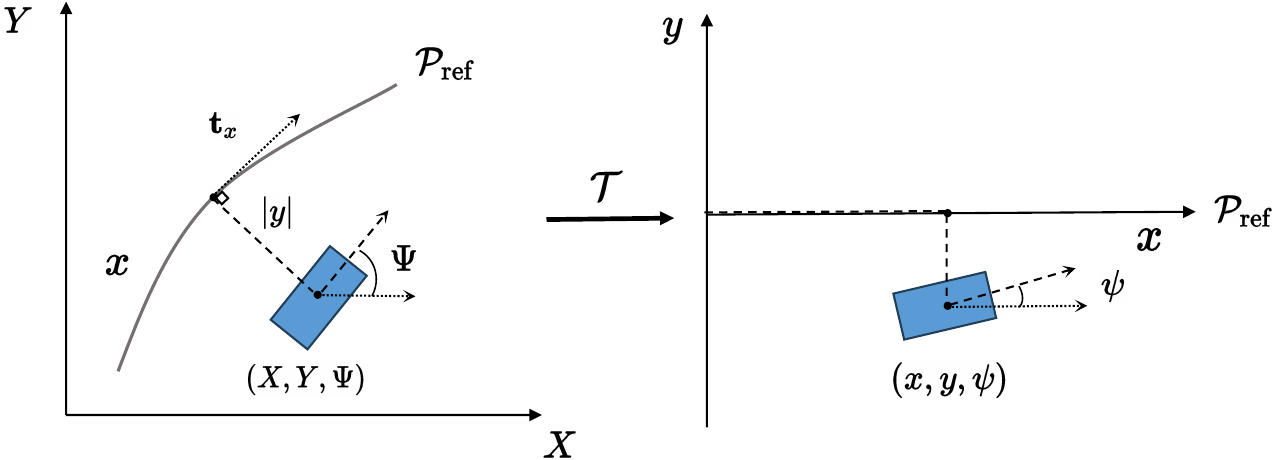}
   \caption{Illustration of the pose transformation $\mathcal{T}$ from the global
coordinate frame $\mathcal{W}_g$ to the centerline reference coordinate frame $\mathcal{W}_\mathrm{ref}$.}
    \label{transform}
\end{figure}
According to \cite{eiras2021two}, 
we can define an invertible transformation $\mathcal{T}$ that maps the pose of the vehicle from the global coordinate $\mathcal{W}_g$ to centerline reference coordinate $\mathcal{W}_\mathrm{ref}$ (as shown in Fig.~\ref{transform})
, which is given by:
\begin{equation}
    (x,y,\psi) = \mathcal{T} (X,Y,\Psi),
\end{equation}
where $x,\ y$, and $\psi$ are the X-coordinate position, Y-coordinate position, and heading angle of the vehicle in the road centerline reference coordinate $\mathcal{W}_\mathrm{ref}$. 
Specifically, we have
\begin{equation}
    x=\underset{\lambda}{\operatorname{argmin}}\left(X-X^{\mathcal{P}_{\mathrm{ref}}}(\lambda)\right)^2+\left(Y-Y^{\mathcal{P}_{\mathrm{ref}}}(\lambda)\right)^2,
\end{equation}
\begin{equation}
    y=\frac{1}{\left\|\mathbf{n}_x\right\|} \mathbf{n}_x^{\top} \cdot \left[\begin{array}{c}X-X^{\mathcal{P}_{\mathrm{ref}}}(x) \\ Y-Y^{\mathcal{P}_{\mathrm{ref}}}(x)\end{array}\right],
\end{equation}
\begin{equation}
    \psi=\Psi - \arctan \left(\left.\frac{\partial Y^{\mathcal{P}_{\mathrm{ref}}}(\lambda)}{\partial X^{\mathcal{P}_{\mathrm{ref}}}(\lambda)}\right|_{\lambda=x}\right).
\end{equation}
Moreover, as $\mathcal{T}$ is a spatial transformation, the speed keeps invariant, i.e., $v=V$.
Thus, the vehicle state vector is redefined as $\mathbf{x} = \begin{bmatrix} x & y & \psi & v \end{bmatrix} ^{\top}$ in the centerline reference coordinate $\mathcal{W}_\mathrm{ref}$.

Urban self-driving is a compound task consisting of overtaking, lane change, and collision avoidance in residential scenarios where other traffic participants (such as other vehicles, motorcyclists, cyclists, and pedestrians) are considered on the road.
The objective of this work is to 
 generate a sequence of optimal states and control commands for the ego vehicle to reach the destination while optimizing driving efficiency and minimizing the risk of collisions with road boundaries and other traffic participants.

Mathematically, with the given sampling time $d_t$ and vehicle model $f$, a sequence of vehicle states $\mathbf{x}_k, \forall k \in\{0, 1, \cdots, N\}$ and control commands $\mathbf{u}_k, \forall k \in\{0, 1, \cdots, N-1\}$ are discretized over a receding horizon $N$. Let $\mathbf{x}_g$ denote the vehicle terminal state of the task, the control objective is to generate the optimal state trajectory $\boldsymbol{\xi}^*=\left\{\mathbf{x}_k^*\right\}_{k=0}^N$ towards $\mathbf{x}_g$ and a sequence of optimal control command $\boldsymbol{\zeta}^*=\left\{\mathbf{u}_k^*\right\}_{k=0}^N$, while increasing the average driving speed $\Bar{v}$ and decreasing the probability of collisions with road boundaries and other traffic participants $p_\mathrm{coll}({\mathbf{x}_k^*})$.

Given the nearest boundary point $\mathcal{B} \in \mathbb{R}^2$ in a drivable surface area, current and predicted state trajectories of other traffic participants which are denoted by $\mathbf{s}_k^i=\left(x_k^i, y_k^i, \psi_k^i, v_k^i\right)$, 
\textcolor{black}{$i \in \mathbb{N}$}, the urban driving task can be formulated as an MPC problem over the receding horizon $N$. It leads to a nonlinear and nonconvex constrained optimization problem:
\begin{subequations}
\begin{align}
&\min _{\mathbf{x}_{1: N}, \mathbf{u}_{0: N-1}} J(\mathbf{x}_k,\mathbf{u}_k,\mathbf{\Delta u}_k) \label{mpc_org_cost}\\
&\text {\quad \quad s.t.} \quad \mathbf{x}_{k+1}=\mathbf{x}_k+f\left(\mathbf{x}_k, \mathbf{u}_k\right) d_t, \label{mpc_org_a}\\
& \quad \quad \quad \quad \mathcal{G}(v_k,\mathbf{u}_k)\leq 0, \label{mpc_org_b}\\
& \quad \quad \quad \quad \mathcal{H}(\mathbf{x}_k,\mathbf{s}_k^i,\mathcal{B})\geq 0,\
\label{mpc_org_c}
\end{align}
where $\mathbf{\Delta u}_k$ is the variation of control commands, $v_k$ is the speed of the ego vehicle, \textcolor{black}{(\ref{mpc_org_cost}) is the cost functions to be optimized}, (\ref{mpc_org_a}) is the vehicle's dynamic constraints, (\ref{mpc_org_b}) represents the box constraints of velocity and control commands and (\ref{mpc_org_c}) is the safety constraints arising from drivable surface boundaries and collision avoidance with other traffic participants.
\label{mpc_org}
\end{subequations}

\section{Learning-based Model Predictive Control}
\label{sec:mpc}
\subsection{Online MPC Reformulation with Instantaneous References}
{Generally, problem (\ref{mpc_org_cost}-\ref{mpc_org_c}) relies on prior knowledge about states of traffic, demanded by external traffic perception and prediction modules or even under the assumptions of perfect scene perception and prediction. Thus, this approach does not maintain the consistency of a unified self-driving framework.}
Moreover, the safety constraints (\ref{mpc_org_c}) typically lead to the nonlinear characteristics and nonconvexity of \textcolor{black}{the above problem}. 
Consequently, solving this problem online usually suffers from a heavy computational burden and poor-quality solutions.

\textcolor{black}{We introduce a learnable state vector $\mathbf{x}_\mathrm{ref}$ to guide the ego vehicle to drive in the drivable and safe region:}
 \begin{equation}
     \mathbf{x}_\mathrm{ref} =
    \begin{bmatrix} x_{\operatorname{ref}}& y_{\operatorname{ref}}& \psi_{\operatorname{ref}}& v_{\operatorname{ref}} \end{bmatrix} ^{\top} \in \mathbb{R}^{4},
 \end{equation}
where $x_{\operatorname{ref}}, y_{\operatorname{ref}}, \psi_{\operatorname{ref}}, v_{\operatorname{ref}}$ represent intermediate reference state variables, including X-coordinate and Y-coordinate positions, heading angle, and speed, which are time-invariant variables over $N$. 

Here, we exploit a neural network policy $\pi$ to reason about the traffic from raw onboard sensor data and infer the intermediate state vector $\mathbf{x}_\mathrm{ref}$.
We further design an instantaneous and self-tuned cost term $J_{\operatorname{ref}, k}$ for driving guidance:

\begin{equation}
    J_{\operatorname{ref}, k}=\left\|\mathbf{x}_k-\mathbf{x}_{\operatorname{ref}}\right\|_{\mathbf{Q}_{\operatorname{ref}}}^2,
\end{equation}
where $\mathbf{Q}_{\operatorname{ref}} = \operatorname{diag}\left(\left[Q_{x_{\operatorname{ref}}}, Q_{y_{\operatorname{ref}}}, Q_{\psi_{\operatorname{ref}}}, Q_{v_{\operatorname{ref}}}\right]\right)$ is \textcolor{black}{also a learnable vector}, which is the time-varying positive semi-definite diagonal weighting matrix to define the relative importance of designed reference cost among all costs.

\textcolor{black}{For $J(\mathbf{x}_k,\mathbf{u}_k,\mathbf{\Delta u}_k)$, we decompose it into four cost terms, 
with target-oriented stage cost $J_{x_k}=\left\|\mathbf{x}_k-\mathbf{x}_g\right\|_{\mathbf{Q}_x}^2$, terminal cost $J_{x_N}=\left\|\mathbf{x}_N-\mathbf{x}_g\right\|_{\mathbf{Q}_x}^2$, energy consumption cost $J_{u_k}=\left\|\mathbf{u}_k\right\|_{\mathbf{Q}_u}^2$, and driving comfort cost $J_{\Delta u_k}=\left\|\mathbf{u}_k - \mathbf{u}_{k-1}\right\|_{\mathbf{Q}_{\Delta u}}^2$,
where $\mathbf{Q}_{x}, \mathbf{Q}_{u},$ and $\mathbf{Q}_{\Delta_u}$ are time-invariant positive semi-definite diagonal matrices.}

With the above settings, the urban driving tasks can be reformulated as the following nonlinear optimization problem {to be} solved online over the receding horizon $N$: 
\begin{equation}
\begin{aligned}
&\min _{\mathbf{x}_{1: N}, \mathbf{u}_{0: N-1}} J_{x_{N}} + \sum_{k=0}^{N-1}\left(J_{x_k}+ J_{u_k}+J_{\Delta u_k}+J_{\operatorname{ref}, k}\right)\\
&\text {\quad \quad s.t.} \quad \mathbf{x}_{k+1}=\mathbf{x}_k+f\left(\mathbf{x}_k, \mathbf{u}_k\right) d_t, \\
 & \quad \quad \quad \quad \mathbf{x}_0=\mathbf{x}_{\text {init }}, \mathbf{u}_{-1}=\mathbf{u}_{\text {init }},\\
 & \quad \quad \quad \quad v_{\min } \leq v_k \leq v_{\max }, \\ 
 & \quad \quad \quad \quad a_{\min } \leq a_k \leq a_{\max }, \\ 
 & \quad \quad \quad \quad -\delta_{\max } \leq \delta_k \leq \delta_{\max },
\label{mpc_re} 
\end{aligned}
\end{equation}
 where the initial states and initial control commands are represented by $\mathbf{x}_{\text {init}}$ and $\mathbf{u}_{\text {init}}$. Also, the speed of the ego vehicle is limited by $v_{\min}$ and $v_{\max}$ according to the traffic rules, and the control commands are constrained by $a_{\min}$, $a_{\max}$, and $-\delta_{\max}$, $\delta_{\max}$ considering the physical limits of vehicle model, respectively. 
\subsection{Real-Time Decision Variable Generation} 

In this work, we integrate $\mathbf{x}_\mathrm{ref}$ and $\mathbf{Q}_{\mathrm{ref}}$ into a decision vector $\mathbf{z}$ as references to reformulate the MPC, which is defined as follows:
 \begin{equation}
    \mathbf{z} = \begin{bmatrix} \mathbf{x}^\top_\mathrm{ref} & {\text{vec}(\mathbf{Q}_{\mathrm{ref}})}^\top \end{bmatrix}^{\top} \in \mathbb{R}^{8}.
 \end{equation}

Let $f_\mathrm{MPC}: \mathbb{R}^{8} \rightarrow \mathbb{R}^{4N}$ denote the mapping function of MPC. 
It is noted that various optimal state trajectories $\boldsymbol{\xi}^*=\left\{\mathbf{x}_k^*\right\}_{k=0}^N \in \mathbb{R}^{4N}$ can be generated by feeding MPC with different decision vectors:
\begin{equation}
    \boldsymbol{\xi}^*(\mathbf{z}) = f_\mathrm{MPC}(\mathbf{z}).
\end{equation}

Therefore, it is imperative to obtain the desired decision variables as instantaneous references to formulate the MPC.
Hence, we can incorporate a step-based DRL technique to learn the optimal neural policy $\pi^*$, such that the decision variables can be determined in real time. 

\section{Learning the Decision Variables via DRL}
\label{sec:rl}




\subsection{Partial Observation}
In the sequel, we use the notations with subscript $t$ to represent their respective values at decision step $t$. 
With $T$ as {the time horizon of episodes}, the generation of decision variables through the inference of high-level policy $\pi$ is a sequential decision-making problem in essence, which can be recorded as:
\begin{equation}
    \mathcal{Z} = (\mathbf{z}_0^\top, \mathbf{z}_1^\top, \dots, \mathbf{z}_{T-1}^\top).
\end{equation}

Considering the online MPC as the low-level planer, at each decision step $t$, the decision variables are determined \textcolor{black}{by reasoning about the traffic from raw onboard sensor data}, as the references to modulate the cost functions of MPC. In this work, the observations of the RL agent at each decision step $t$ take the partial form as $\mathbf{o}_t = (\tilde{x}_t, y_t, \psi_t, v_t, \boldsymbol{d}_t)$, and it is defined in Table \ref{obs_tab}. 
In this table, $\tilde{x}_t$ is the distance of the X-coordinate position between current vehicle state $\mathbf{x}_t$ and terminal state $\mathbf{x}_g$ in centerline reference coordinate $\mathcal{W}_\mathrm{ref}$. $y_t$, $\psi_t$, and $v_t$ are the Y-coordinate position, heading angle, and speed of $\mathbf{x}_t$ in $\mathcal{W}_\mathrm{ref}$, respectively.  
\begin{table}[h]
    \centering
    \caption{Observation Space ($\mathbb{R}^{4+n}$)}\label{obs_tab}
    \begin{tabular}{ccc}
        \toprule
        $\tilde{x}_t$ & Distance to terminal state of X-position in $\mathcal{W}_\mathrm{ref}$ & $\mathbb{R}$  \\
        $y_t$  &  Y-position in $\mathcal{W}_\mathrm{ref}$ & $\mathbb{R}$ \\
        $\psi_t$  &  Heading angle in $\mathcal{W}_\mathrm{ref}$ & $\mathbb{R}$ \\
        $v_t$   &  Speed in $\mathcal{W}_\mathrm{ref}$ & $\mathbb{R}$\\
        $\boldsymbol{d}_t$   &  2D Lidar distance measurements ($-90^{\circ}, 90^{\circ}$, $\SI{50}{m}$) & $\mathbb{R}^{n}$ \\
        \bottomrule
    \end{tabular}
\end{table}

Essentially, we utilize a practical and lightweight method that makes use of an $n$-line 2D Lidar to collect the raw sensor data to perceive the surrounding traffic. 
Moreover, we apply $z$-score normalization to whitening the observation features for stable and efficient training.


\subsection{Action} 

At each decision step $t$, given the agent’s observation $\mathbf{o}_t$, the policy network outputs continuous action $\mathbf{a}_t$ as the decision vector $\mathbf{z}_t$, which is denoted as: 
\begin{equation}
    \mathbf{z}_t = \mathbf{a}_t = \pi(\mathbf{o}_t).
\end{equation}

The action space of RL is set to be continuous, whose lower and upper bounds are given in Table \ref{val_range}.
\begin{table}[h]
    \centering
    \caption{Value Ranges of Decision Variables}\label{val_range}
    \begin{tabular}{cccc}
        \toprule
        Variable & Interval & Variable & Interval\\
        $x_{\operatorname{ref},t}$  &  $[-40,20]$ &  $Q_{x_{\operatorname{ref},t}}$  &  $[0,50]$ \\
        $y_{\operatorname{ref},t}$  &  $[-15,15]$ &  $Q_{y_{\operatorname{ref},t}}$  &  $[0,50]$ \\
        $\psi_{\operatorname{ref},t}$   &  $[-\pi/2,\pi/2]$ &  $Q_{\psi_{\operatorname{ref},t}}$   &  $[0,50]$ \\
        $v_{\operatorname{ref},t}$   &  $[-10,20]$  &  $Q_{v_{\operatorname{ref},t}}$   &  $[0,50]$  \\
        \bottomrule
    \end{tabular}
\end{table}

\begin{algorithm}[t]  
	\caption{Online Motion Planning and Control with MPC Through Policy Inference
 } \label{algo-inference}
	\LinesNumbered 
	\KwIn{$f_\mathrm{MPC}$, $\mathbf{x}_g$}
	\KwOut{$u_{\operatorname{th},t}, u_{\operatorname{br},t}, u_{\operatorname{st},t}$}
        
        Sample $\mathbf{o}_t$\;
        Normalize $\mathbf{o}_t$\;
        $\mathbf{z}_t = \mathbf{a}_t = \pi^*(\mathbf{o}_t)$\;
        $J_{\operatorname{ref}, k}=\left\|\mathbf{x}_k-\mathbf{x}_{\operatorname{ref}}\right\|_{\mathbf{Q}_{\operatorname{ref}}}^2$\;
        Solve (\ref{mpc_re}) online to get $\boldsymbol{\xi}^*(\mathbf{z}_t)$ and $\boldsymbol{\zeta}^*(\mathbf{z}_t)$\;
        $(u_{\operatorname{th},t}, u_{\operatorname{br},t}, u_{\operatorname{st},t}) = f_{\mathrm{c}}(\mathbf{u}_t)$\;
        
\end{algorithm}

It is noted that $x_{\operatorname{ref},t}$ is considered as the deviation of the current X-coordinate position, i.e., $\tilde{x}_{\operatorname{ref},t} \leftarrow x_{\operatorname{ref},t} + x_t$, and $\mathbf{Q}_{\mathrm{ref},t}$ takes the proportion form of $\mathbf{Q}_{x}$, i.e., $\tilde{\mathbf{Q}}_{\mathrm{ref},t}
\leftarrow \mathbf{Q}_{\mathrm{ref},t} \odot \mathbf{Q}_{x}$, where $\odot$ is the Hadamard product for  element-wise multiplication.
With the generated step actions $\mathbf{a}_t$ as the decision vector $\mathbf{z}_t$, the corresponding MPC is modulated to generate a sequence of optimal state trajectories $\boldsymbol{\xi}^*(\mathbf{z}_t)=\left\{\mathbf{x}_k^*\right\}_{k=0}^N$  and control command $\boldsymbol{\zeta}^*(\mathbf{z}_t)=\left\{\mathbf{u}_k^*\right\}_{k=0}^N$. Finally, the first tuple $\mathbf{u}_0^*$ in $\left\{\mathbf{u}_k^*\right\}_{k=0}^N$ is converted to control signals of throttle $u_{\operatorname{th},t}$, brake $u_{\operatorname{br},t}$, and steering angle $u_{\operatorname{st},t}$ by a command converter $f_{\mathrm{c}}$ to transit the vehicle state. 
The command converter $f_{\mathrm{c}}$ is defined as
$
    (u_{\operatorname{th},t}, u_{\operatorname{br},t}, u_{\operatorname{st},t}) = f_{\mathrm{c}}(\mathbf{u}_t),
$
where
 \begin{align*}
    & (u_{\operatorname{th},t}, u_{\operatorname{br},t}) = \begin{cases} (\min(a_t/3, 1),0), & a_t \geq 0 \\ (0, \max(-a_t/8, -1)), & a_t < 0\end{cases},\\
    & u_{\mathrm{st},t} = \textup{clip}(\delta_t, -1, 1).
 \end{align*}
Note that \textup{clip($\cdot$)} is a clip function to prevent values from exceeding the prescribed threshold $[-1,1]$.




The motion planning and control procedures are executed according to Algorithm \ref{algo-inference}.

\subsection{Reward}

To encourage safe and efficient self-driving behaviors, after taking each decision $\mathbf{z}_t = \mathbf{a}_t$, the RL agent receives a reward signal $r_t$ with the reward function:
\begin{equation}
    r_t(\mathbf{a}_t | \mathbf{o}_t) = r_\mathrm{forward} + r_\mathrm{speed} + r_\mathrm{coll} + r_\mathrm{road} + r_\mathrm{steer} + r_\mathrm{time}.
\end{equation}

More details of the rewards are shown in Table \ref{reward_tab}, where $y_\mathrm{{bound}}$ is the Y-coordinate position of the nearest road boundary in $\mathcal{W}_\mathrm{ref}$.

\vspace{-0.3cm}
\begin{table}[h]
    \centering
    \caption{Reward Function}\label{reward_tab}
    \begin{tabular}{ccc}
        \toprule
        $r_\mathrm{forward}$  &  Step forward distance & $\tilde{x}_{t-1}-\tilde{x}_{t}$\\ $r_\mathrm{speed}$   &  Average speed if reaching destination & $\Bar{v}$\\
        $r_\mathrm{coll}$ & Penalty if collision occurs & -100  \\
        $r_\mathrm{road}$   &  Deviation distance if out of road & $-|y_t - y_\mathrm{{bound}}|$\\
        $r_\mathrm{steer}$   &  Steering cost & $-|\delta|$\\
        $r_\mathrm{time}$  &  Punishment if out of time & -100 \\
        \bottomrule
    \end{tabular}
\end{table}

\subsection{Policy Training}

The corresponding transition $\left(\mathbf{o}_t, \mathbf{a}_t, r_t, \mathbf{o}_{t+1}\right)$ including observations, actions, and rewards are stored into the fixed-size first-in, first-out replay buffer $\mathcal{D}$ for offline training.
We utilize the off-policy SAC algorithm \cite{haarnoja2018soft, haarnoja2018softapp} to learn the optimal policy $\pi^*$ that maximizes the 
return together with its entropy:
\begin{equation}
\pi^*=\operatorname{argmax} \mathbb{E}_{\left(\mathbf{o}_t, \mathbf{a}_t\right)}\left[\sum_{t=0}^T \gamma^t\left(r_t+\alpha \mathcal{H}\left(\pi\left(. \mid \mathbf{o}_t\right)\right)\right)\right],
\end{equation}
where $\gamma$ is the discount factor, $\mathcal{H}$ denotes the entropy, and $\alpha$ is the temperature parameter that tunes the importance of the entropy term versus the return.

To avoid the critic learning oscillation and subsequent performance deterioration, the conditions of terminating the episode are to be distinguished before entering the replay buffer, i.e., we set the value to 0 if and only if a collision occurs. Moreover, the disparity in magnitude between the reward values can lead to a sudden change in the Q-function, which hampers the learning process. Thus, 
we \textcolor{black}{reshape the reward once an outlier reward is given, which significantly improves the effectiveness of training:}
\begin{equation}
r_t= \begin{cases}-5, & r_t \leq -5, \\ r_t, & \text { otherwise. }\end{cases}
\end{equation}

The policy training process is summarized in Algorithm~\ref{algo-learning}.

\begin{algorithm}[t]  
	\caption{Policy Training with Off-Policy RL} \label{algo-learning}
	\LinesNumbered 
	\KwIn{$f_\mathrm{MPC}$}
	\KwOut{$\pi^*_{\boldsymbol{\theta}}$}
        Initialize $\boldsymbol{\theta}$ and empty $\mathcal{D}$\; 
        Reset the environment to get $\mathbf{x}_g$ and $\mathbf{o}_t$\;
        \While{not terminated}  
            {Normalize $\mathbf{o}_t$\;
            \uIf{random exploration}
                {Randomly sample $\mathbf{z}_t = \mathbf{a}_t = \text{random()}$\;}
            \Else{$\mathbf{z}_t = \mathbf{a}_t = \pi_{\boldsymbol{\theta}}(\mathbf{o}_t)$\;}
            Solve (\ref{mpc_re}) online to obtain $\boldsymbol{\xi}^*\left(\mathbf{z}_t\right)$ and $\boldsymbol{\zeta}^*(\mathbf{z}_t)$\;
            $(u_{\operatorname{th},t}, u_{\operatorname{br},t}, u_{\operatorname{st},t}) = f_{\mathrm{c}}(\mathbf{u}_t)$\;
            Sample $r_t$, $\mathbf{o}_{t+1}$\;
            Store $\left(\mathbf{o}_t, \mathbf{a}_t, r_t, \mathbf{o}_{t+1}\right)$ in $\mathcal{D}$\;
            \uIf{update policy}
                {Update $\pi_{\boldsymbol{\theta}}$ with SAC\;}}   
\end{algorithm}

\section{Experiments}
\label{sec:exp}

\subsection{Implementation Setup}

The MPC problem is solved using CasADi \cite{andersson2019casadi} with the IPOPT option via the single-shooting method. 
The weighting matrices $\mathbf{Q}_{x}$, $\mathbf{Q}_{u}$, and $\mathbf{Q}_{\Delta_u}$ are set to $\operatorname{diag}\left(\left[100, 100, 100, 10\right]\right)$, $\operatorname{diag}\left(\left[1, 1\right]\right)$, and $\operatorname{diag}\left(\left[0.1, 0.1\right]\right)$. We take the length of receding horizon $N$ as $\SI{5.0}{s}$ and discrete sampling time $d_t$ as $\SI{0.1}{s}$. 
The lower and upper bounds of speed, acceleration, and steering angle are set to $v_{\min} = \SI{0.0}{m/s}$, $v_{\max} = \SI{10.0}{m/s}$, $a_{\min} = \SI{-9.0}{m/s^2}$, $a_{\max} = \SI{4.5}{m/s^2}$, $\delta_{\min} = \SI{-0.75}{rad}$, and $\delta_{\max} = \SI{0.75}{rad} $, respectively. 
We construct the policy and critic networks in PyTorch \cite{paszke2019pytorch}, each with 2
hidden layers with 256 LeakyReLU nodes. The networks are trained with Adam optimizer \cite{kingma2014adam} with a learning rate $3 \times 10^{-4}$. The discount factor is set to $\gamma=0.99$. Training starts after the replay buffer collects more than
2500 randomly explored step data.




The long outer ring road with 3 lanes in Town05 of the high-fidelity simulator CARLA \cite{dosovitskiy2017carla} is the testbed for SAC model training and performance evaluation. 
Here, we set the Tesla Model 3 as the self-driving vehicle and consider 9 traffic participants in various types, e.g., vans, cars, motorcyclists, and cyclists. All other traffic participants are in built-in auto-pilot mode and initialized in random positions. Here, we take a 73-line 2D Lidar for perception, i.e., $n=73$. 

\begin{figure}[htbp]
\centering
\subfigure[Collision avoidance]{
\begin{minipage}[t]{0.31\linewidth}
\centering
\includegraphics[width=2.6cm, height=2.4cm]{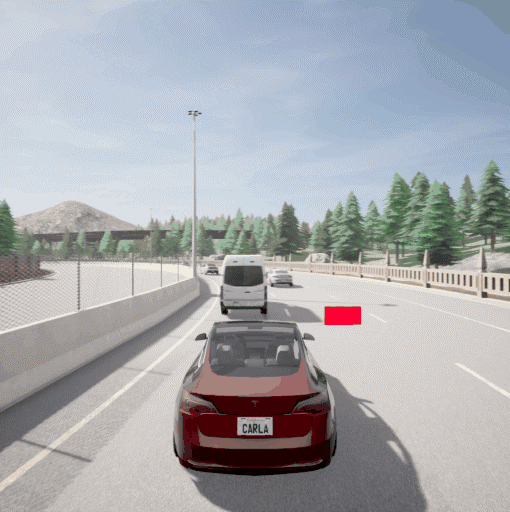}
\end{minipage}%
}%
\subfigure[Lane change]{
\begin{minipage}[t]{0.31\linewidth}
\centering
\includegraphics[width=2.6cm, height=2.4cm]{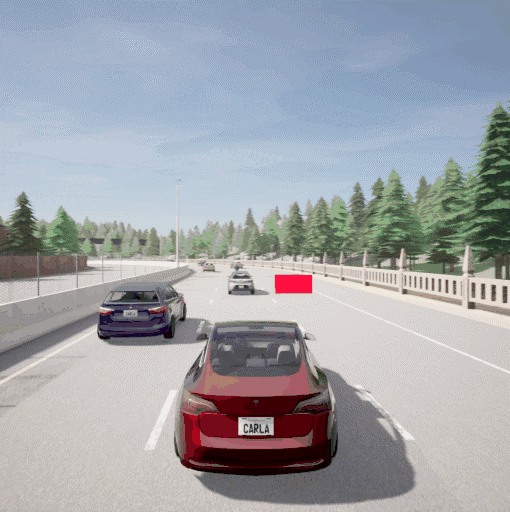}
\end{minipage}%
}%
\centering
\subfigure[Overtaking]{
\begin{minipage}[t]{0.31\linewidth}
\centering
\includegraphics[width=2.6cm, height=2.4cm]{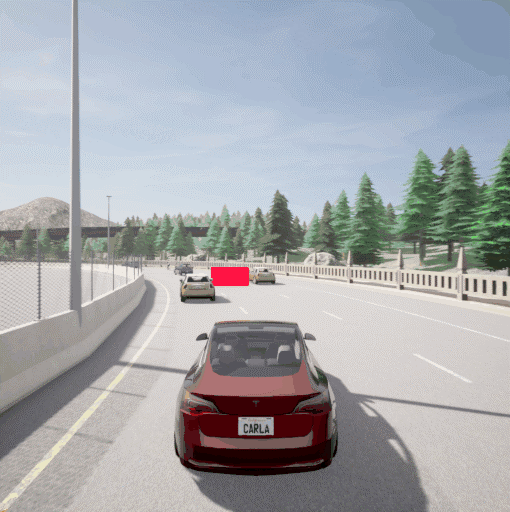}
\end{minipage}%
}%
\centering
\caption{Visualization of learned positions in decision variables for different driving scenarios.}
\vspace{-0.5cm}
\label{visualization}
\end{figure}

\begin{figure*}[htbp]
\centering
\subfigure[$t=\SI{17.0}{s}$]{
\begin{minipage}[t]{0.24\linewidth}
\centering
\includegraphics[width=4.4cm, height=2.2cm]{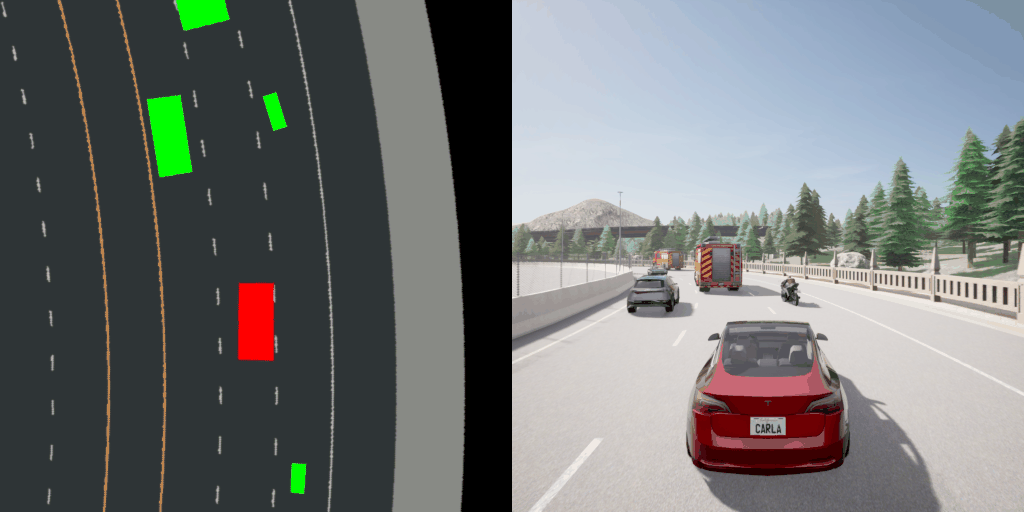}
\end{minipage}%
}%
\subfigure[$t=\SI{18.3}{s}$]{
\begin{minipage}[t]{0.24\linewidth}
\centering
\includegraphics[width=4.4cm, height=2.2cm]{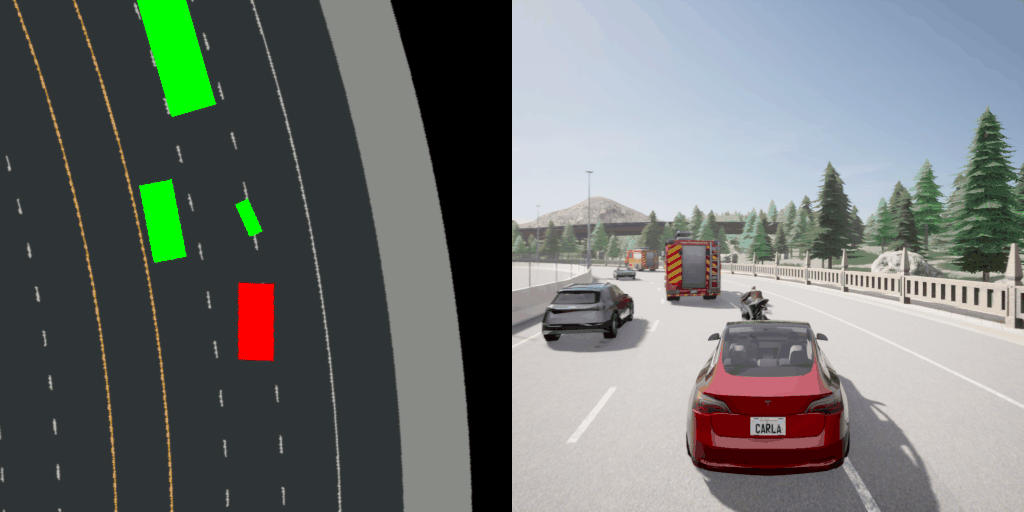}
\end{minipage}%
}%
\centering
\subfigure[$t=\SI{19.0}{s}$]{
\begin{minipage}[t]{0.24\linewidth}
\centering
\includegraphics[width=4.4cm, height=2.2cm]{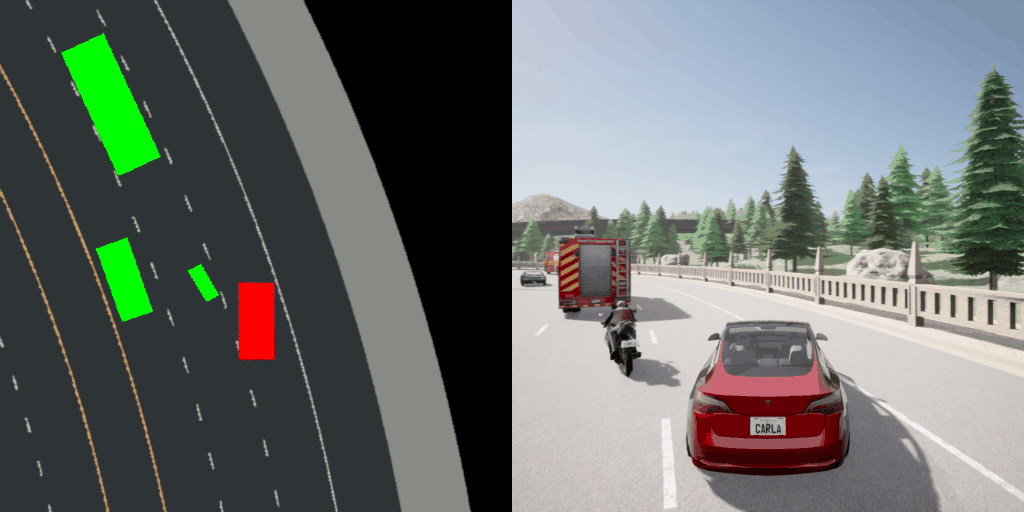}
\end{minipage}%
}%
\centering
\subfigure[$t=\SI{24.5}{s}$]{
\begin{minipage}[t]{0.24\linewidth}
\centering
\includegraphics[width=4.4cm, height=2.2cm]{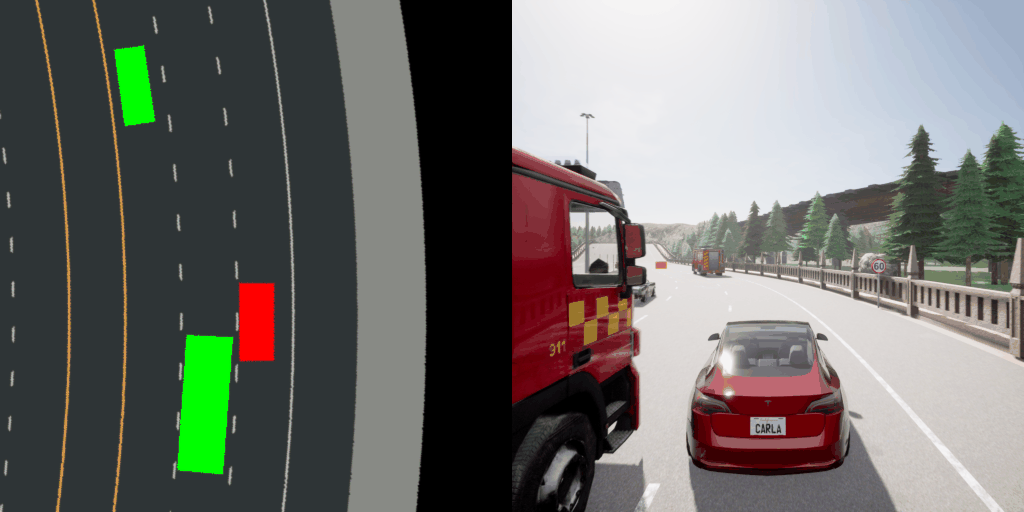}
\end{minipage}%
}%
\centering

\centering
\subfigure[$t=\SI{11.4}{s}$]{
\begin{minipage}[t]{0.24\linewidth}
\centering
\includegraphics[width=4.4cm, height=2.2cm]{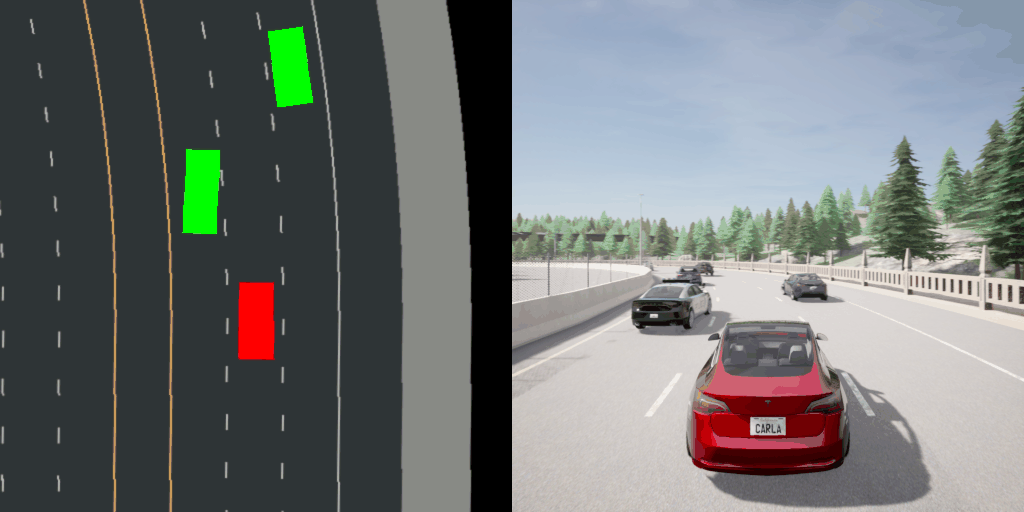}
\end{minipage}%
}%
\subfigure[$t=\SI{12.2}{s}$]{
\begin{minipage}[t]{0.24\linewidth}
\centering
\includegraphics[width=4.4cm, height=2.2cm]{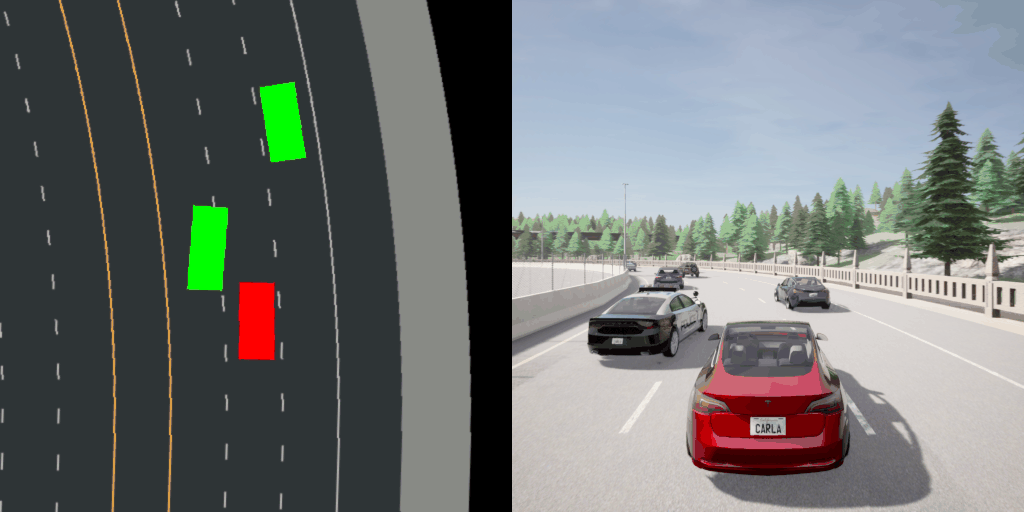}
\end{minipage}%
}%
\centering
\subfigure[$t=\SI{13.2}{s}$]{
\begin{minipage}[t]{0.24\linewidth}
\centering
\includegraphics[width=4.4cm, height=2.2cm]{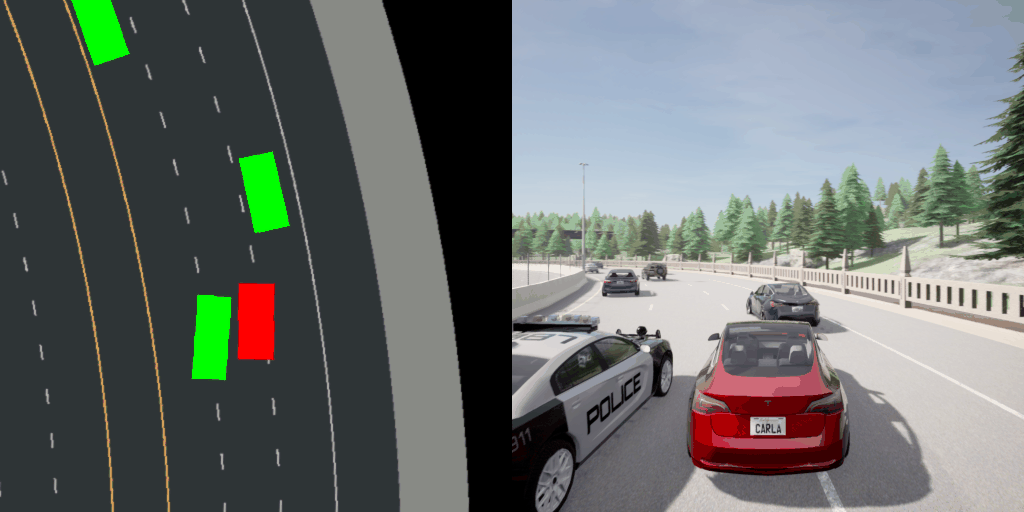}
\end{minipage}%
}%
\centering
\subfigure[$t=\SI{14.0}{s}$]{
\begin{minipage}[t]{0.24\linewidth}
\centering
\includegraphics[width=4.4cm, height=2.2cm]{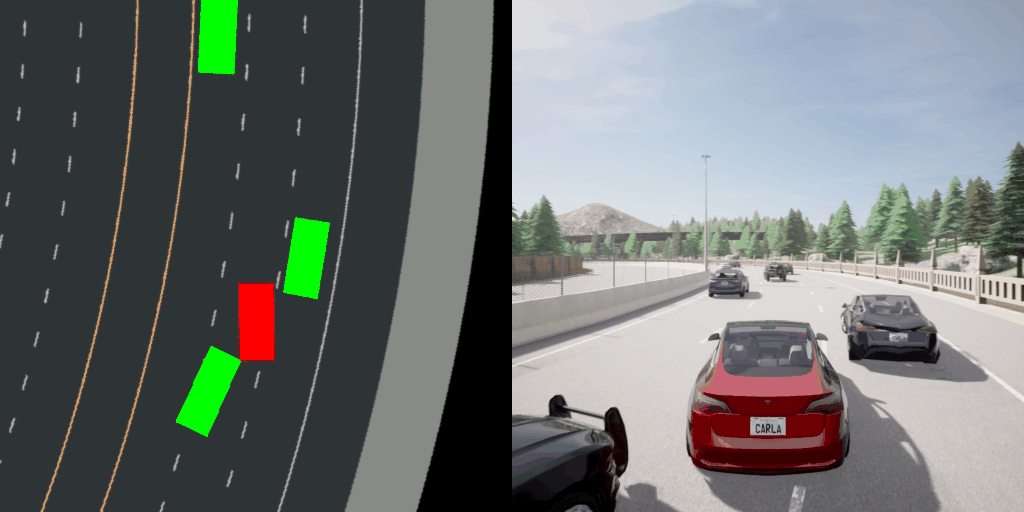}
\end{minipage}%
}%
\centering

\caption{Key frames of trails with our framework in complex and dynamic traffic environments. 
\textcolor{black}{\textbf{Top: }agile overtaking. \textbf{Bottom:} emergent collision avoidance.}
The left side of each subfigure is a bird-view image where the red rectangle is the ego vehicle and the green rectangles represent other traffic participants.}
\label{agility}
\vspace{-0.4cm}
\end{figure*}

\subsection{Driving Performance}
\textbf{1) Interpretability of Safety:} \textcolor{black}{The interpretability of driving safety of our framework is investigated and visualized in Fig. \ref{visualization}}, where we highlight the learned reference positions in $\mathbf{x}_\mathrm{ref}$ using red waypoints. \textcolor{black}{Our approach is able to reason about the real-time traffic and decide the reference positions for dealing with both the short-term potential collision avoidance (see Fig. \ref{visualization}(a)) and long-term planning such as lane change and overtaking (see Fig. \ref{visualization}(b, c)).}

\textbf{2) Maneuverability:}
 Examples are displayed to illustrate the superiority in terms of driving maneuverability as shown in Fig. \ref{agility}. It is noted in the top of Fig. \ref{agility} that the current traffic is a complex dilemma at $t=\SI{17.0}{s}$, as all lanes are occupied with traffic participants in driving. 
 The motorcyclist on the right lane intends to change into the middle lane at $t=\SI{18.3}{s}$, leaving the right lane temporarily empty. In this case, the only way to escape from this dilemma is to traverse the temporarily empty lane to enter the open road; otherwise, the ego vehicle would sacrifice driving efficiency to follow the front vehicles. 
 At $t=\SI{19.0}{s}$, the ego vehicle turns right to occupy the temporarily empty lane and then accelerates to overtake the surrounding vehicles. Finally, the ego vehicle enters the open road and escapes from the traffic dilemma at $t=\SI{24.5}{s}$. 

We also evaluate the maneuverability when handling with emergent potential collisions caused by traffic uncertainties. As illustrated in the bottom of Fig. \ref{agility}, the ego vehicle attempts to overtake the two front vehicles by traversing the middle lane at $t=\SI{11.4}{s}$. However, we can find that the front vehicle on the left lane intends to cut in and block the on-taking route of our ego vehicle at $t=\SI{12.2}{s}$. Then, the ego vehicle urgently dodges and attempts to bypass the left vehicle with enough safe distance margin at $t=\SI{13.2}{s}$. Ultimately, the emergence of the potential collision caused by unpredictable lane change behavior of the front vehicle is released $t=\SI{14.0}{s}$. 
Therefore, the superiority in maneuverability of our framework for motion planning and control in complex and dynamic traffic is clearly demonstrated.

\subsection{Comparison Analysis}
We compare our proposed framework with the following representative baseline methods:
\begin{itemize}
\item Vanilla-RL: The MPC is removed from the proposed framework, where the policy network directly encodes the input features to actions as control commands \cite{fuchs2021super}. 
    \item Hard-MPC: 
    The hard constraints of collision avoidance with other traffic participants and drivable surface boundaries are implemented according to \cite{eiras2021two}, 
     \textcolor{black}{except that the warm start for solving the optimization is removed for  fairness of comparison.} 
    \item Soft-MPC: The aforementioned constraints are further formulated in the soft form, i.e., the penalty terms are added in the cost functions to limit the amount of constraint violation
    \cite{luis2020online}. 
\end{itemize}

\textcolor{black}{We take simplified traffic for quantitative experiments, where the number of other traffic participants is decreased from 9 to 6 and all traffic participants are uniformed to the Tesla Model 3. As Hard-MPC and Soft-MPC require the predicted states of surrounding participants to implement the safety constraints, a naive motion prediction is adopted, where the participants are assumed to keep the current speed to drive forward along the current heading angle. Besides, oracle traffic states are also assumed for the aforementioned conventional methods.}

We run 100 trials of different methods in the simplified traffic and record their performance in terms of
success, collision and time-out rate, average driving speed as well as computation time of solving the optimization 
as shown in Table \ref{results_tab}. 
\begin{table*}[t]
    \centering
    \caption{Performance in urban driving with different methods }\label{results_tab}
    \begin{tabular}{cccccccc}
        \toprule
        Approaches &  Oracle & Traff. Pred. & Succ. (\%) & Coll. (\%) & Time-out (\%)&  Aver. speed ($\mathrm{m/s}$) & Comp. time ($\mathrm{s}$)\\
        \midrule
        \textbf{Ours}  & - & - & $\bm{90}$ & $\bm{10}$ & $\bm{0}$ & $\bm{8.36}$ & $\bm{0.038}$ \\
            Vanilla-RL  & - & - & 74 & 26 & 0 & 8.25 & - \\ 
            Hard-MPC  & \checkmark & \checkmark  & 64 & 24 & 12 & 5.32 & 0.132 \\
            Soft-MPC  & \checkmark & \checkmark  & 75 & 25 & 0 & 5.97 & 0.076 \\
        \bottomrule
    \end{tabular}
    \vspace{-0.4cm}
\end{table*}
We observe from the table that our method reaches the highest success rate of $90\%$ and the lowest collision rate of $10\%$ with a time-out rate of $0\%$, while obtaining the fastest average driving speed at $\SI{8.36}{m/s}$ \textcolor{black}{and the minimum of solving time of $\SI{0.038}{s}$} compared to the baselines. It is clearly demonstrated that our proposed framework manifests superiority in terms of safety and driving efficiency, \textcolor{black}{as well as alleviated computational burden of solving the optimization}. The improved safety can be elaborated as the safety conditions are encoded by our latent representation. The high driving efficiency can be expounded by the improved maneuverability, which is supported by the fact that our MPC reorganization reduces the difficulty in computing an efficiency-friendly solution of MPC under such a complex and dynamic environment. 

Vanilla-RL manifests acceptable driving efficiency with an average driving speed of $\SI{8.25}{m/s}$, but the safety is not as strong as ours due to the collision rate of $26\%$. This indicates that the black-box characteristic of the learned policy hampers the stability of such a strategy. Furthermore, this is also a part of our intentions to bridge RL and MPC to improve the stability and safety of learned policy.

We also note that Hard-MPC and Soft-MPC suffer from driving efficiency with the average speed of $\SI{5.32}{m/s}$ and $\SI{5.97}{m/s}$ because the solutions of the MPC problem are too conservative for maneuvers to generate agile driving behaviors for high driving efficiency (i.e., overtaking). 
Specifically, due to the heavy computational burden of solving the nonlinear, nonconvex, and constrained MPC online, the time-delay of control commands of Hard-MPC seriously hinders the real-time driving performance. Intuitively, unacceptable time-delay can lead to the failure of collision avoidance and violation of the limitation of road boundaries. Hence, the safety of Hard-MPC is worse than expected with a collision rate of $24\%$. The other disadvantage of Hard-MPC in terms of the time-out rate of $12\%$ can be interpreted by the predicament that occurs when the initial state of MPC is out of the target state set, i.e., the vehicle will be stuck in this predicament once out of the road. 
Furthermore, oscillation and divergence of solutions also impair the performance of Soft-MPC when the optimization becomes overly complex to solve.
Therefore, it is explanatory that Soft-MPC has a collision rate of $25\%$. 

\subsection{Robustness}
\textcolor{black}{In this section, we analyze the robustness of our framework by transferring the trained models to different testing environments.}

\begin{figure}[htbp]
\centering
\subfigure[$t=\SI{2.7}{s}$]{
\begin{minipage}[t]{0.31\linewidth}
\centering
\includegraphics[width=2.6cm, height=2.4cm]{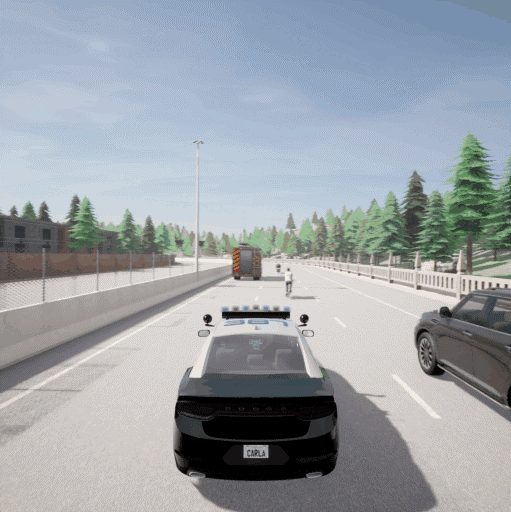}
\end{minipage}%
}%
\subfigure[$t=\SI{4.6}{s}$]{
\begin{minipage}[t]{0.31\linewidth}
\centering
\includegraphics[width=2.6cm, height=2.4cm]{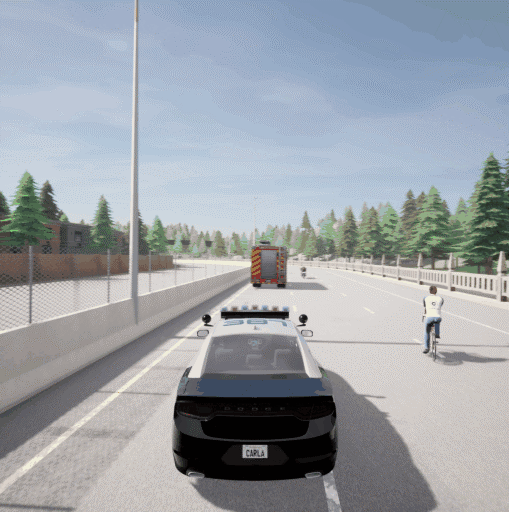}
\end{minipage}%
}%
\centering
\subfigure[$t=\SI{11.2}{s}$]{
\begin{minipage}[t]{0.31\linewidth}
\centering
\includegraphics[width=2.6cm, height=2.4cm]{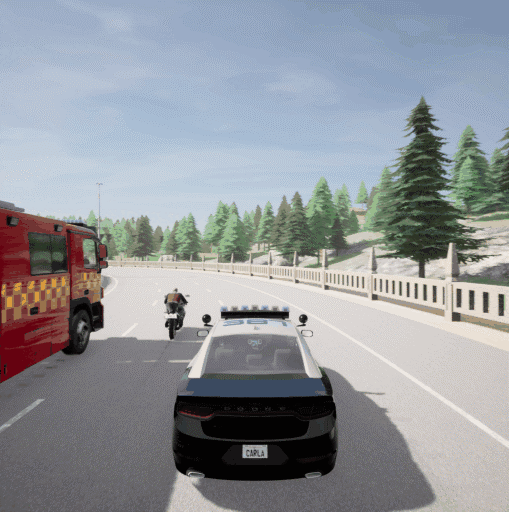}
\end{minipage}%
}%
\centering
\caption{\textcolor{black}{Key frames of a trail of transferring the trained model to a new type of vehicle.}}
\label{transfer}
\end{figure}

\textbf{1) Transfer to new vehicles:} 
\textcolor{black}{We transfer our trained models to Police Charger and evaluate the driving performance. As the trail demonstrated in Fig. \ref{transfer}, we observe that the controlled vehicle still obtains satisfying performance in driving safety for lane change and overtaking tasks.  However, the driving stability and maneuverability are slightly degraded in complex scenarios where more agile maneuvers are required, which are most likely caused by the difference in model shapes and dynamics between the training and testing vehicles. }

\textbf{2) Noisy observations:} 
\textcolor{black}{Table \ref{results_noise} presents the success rate and corresponding loss for varying degrees of uniform noise added to the observations, in comparison to noise-free performance. When exposed to 2\% or 5\% noise, the method demonstrates the performance closely resembling the noise-free one in terms of safety and continues to surpass other baselines. However, increased noise levels in the observations make it challenging to identify feasible and safe references, consequently leading to a decrease in the success rate. With up to 10\% noise, the success rate aligns with the performance of several baselines, indicating a noticeable deterioration in safety.
}
\vspace{-0.4cm}
\begin{table}[h]
    \centering
    \caption{Success rate and its loss of our method under different uniform ranges of noise in observations }\label{results_noise}
    \begin{tabular}{ccc}
        \toprule
        Noise range [\%]  &  Succ. (\%) & Loss. (\%)  \\
        \midrule
            0  & 90 & -  \\
            2  & 86 & 4 \\ 
            5  & 82 & 8   \\
            10  & 75 & 25   \\
        \bottomrule
    \end{tabular}
    \vspace{-0.4cm}
\end{table}

\begin{figure}[t]
	\centering
	\subfigure[$t=\SI{0}{s}$]{
		\begin{minipage}[t]{0.23\linewidth}
		\includegraphics[width=\linewidth]{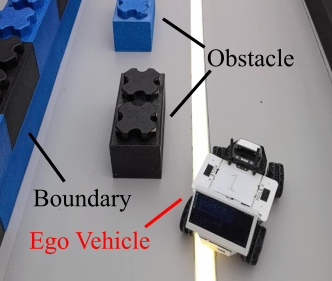}
		\end{minipage}%
	}%
	\centering
	\subfigure[$t=\SI{18.2}{s}$]{
		\begin{minipage}[t]{0.23\linewidth}
		\includegraphics[width=\linewidth]{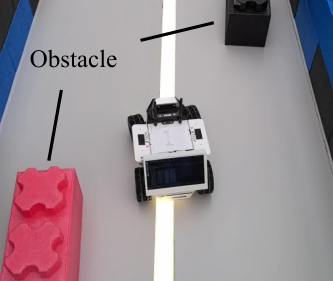}
		\end{minipage}
	}%
	\centering
	\subfigure[$t=\SI{44.6}{s}$]{
		\begin{minipage}[t]{0.23\linewidth}
		\includegraphics[width=\linewidth]{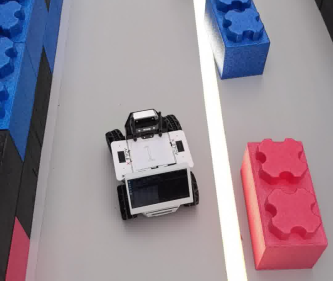}
		\end{minipage}%
	}%
	\centering
	\subfigure[$t=\SI{55.1}{s}$]{
		\begin{minipage}[t]{0.23\linewidth}
		\includegraphics[width=\linewidth]{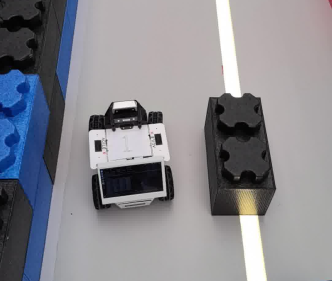}
		\end{minipage}
	}%
 
        \centering
	\subfigure[$t=\SI{11.0}{s}$]{
		\begin{minipage}[t]{0.23\linewidth}
		\includegraphics[width=\linewidth]{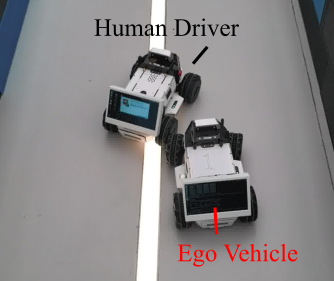}
		\end{minipage}%
	}%
	\centering
	\subfigure[$t=\SI{37.2}{s}$]{
		\begin{minipage}[t]{0.23\linewidth}
		\includegraphics[width=\linewidth]{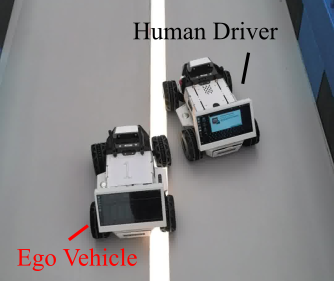}
		\end{minipage}
	}%
	\centering
	\subfigure[$t=\SI{56.9}{s}$]{
		\begin{minipage}[t]{0.23\linewidth}
		\includegraphics[width=\linewidth]{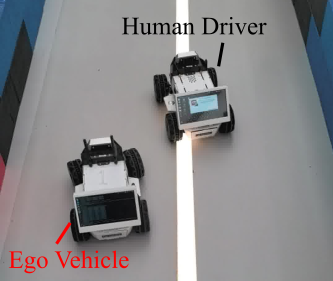}
		\end{minipage}%
	}%
	\centering
	\subfigure[$t=\SI{73.4}{s}$]{
		\begin{minipage}[t]{0.23\linewidth}
		\includegraphics[width=\linewidth]{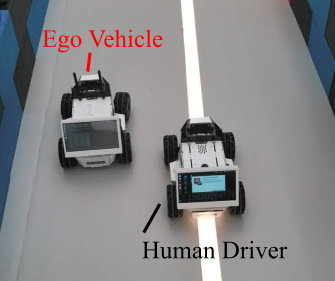}
		\end{minipage}
	}%

        \centering
	\subfigure[$t=\SI{13.0}{s}$]{
		\begin{minipage}[t]{0.23\linewidth}
		\includegraphics[width=\linewidth]{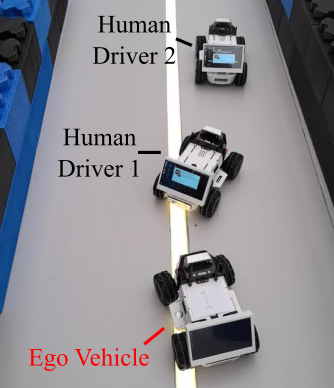}
		\end{minipage}%
	}%
	\centering
	\subfigure[$t=\SI{42.3}{s}$]{
		\begin{minipage}[t]{0.23\linewidth}
		\includegraphics[width=\linewidth]{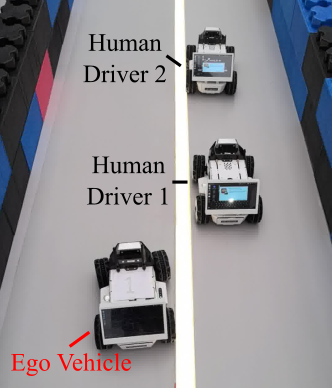}
		\end{minipage}
	}%
	\centering
	\subfigure[$t=\SI{58.5}{s}$]{
		\begin{minipage}[t]{0.23\linewidth}
		\includegraphics[width=\linewidth]{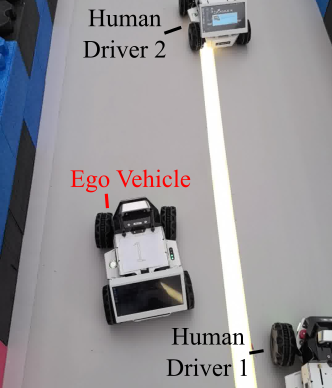}
		\end{minipage}%
	}%
	\centering
	\subfigure[$t=\SI{79.1}{s}$]{
		\begin{minipage}[t]{0.23\linewidth}
		\includegraphics[width=\linewidth]{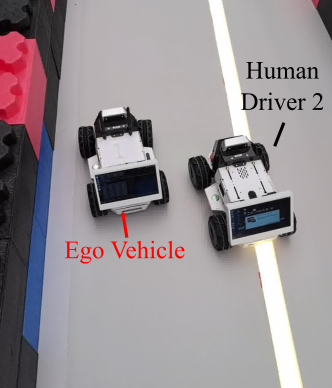}
		\end{minipage}
	}%
 	\caption{Snapshots of the real-world experiments in different scenarios. {\textbf{Top:} lane change in static traffic. \textbf{Middle:} interactive overtaking with a single human driver. \textbf{Bottom:} interactive overtaking with multiple human drivers.}}
 	\label{fig:sim2real}
	\centering
 \vspace{-0.3cm}
\end{figure}

\subsection{{Zero-Shot} Sim-to-Real Transfer}

\textcolor{black}{We deploy the proposed method in real robotic platforms to further evaluate the {zero-shot} generalizability of the presented framework in real-world applications. We use LIMO equipped with the EAI X2L single-line spinning Lidar and NVIDIA Maxwell as our real-world testbed. 
 We design three different driving scenarios to test the performance: lane change in static traffic, interactive overtaking with a single human driver, and interactive overtaking with multiple human drivers. To ensure consistent observations between real world and simulations for smoother sim-to-real transfer, we align the amplitude of observations. As demonstrated in Fig. \ref{fig:sim2real}, the proposed framework manifests remarkable capability in reasoning about static or dynamic traffic with raw sensor data and generating effective collision avoidance and overtaking maneuvers.}

\section{Conclusions}
\label{sec:conc}
In this paper, we propose a novel learning-based MPC framework to address self-driving tasks in complex and dynamic urban scenarios. By modulating the cost functions with instantaneous references in the form of decision variables, the driving safety are latently encoded {without the prior knowledge of oracle and predicted traffic states}. Then, the policy search for real-time generation of desired references is formulated as a DRL problem, where the step actions are cast as the decision variables. Through comprehensive evaluations in a high-fidelity simulator and real-world experiments, our framework manifests superiority among the aspects of driving efficiency, safety, and improved maneuverability.
Our future work is to further enhance the scalability, generalizability and robustness of our method when deploying to other complex and dynamic self-driving scenarios.



\bibliographystyle{IEEEtran}
\bibliography{ref.bib}
\end{document}